\documentclass[runningheads]{llncs}
\usepackage{graphicx}
\usepackage{comment}
\usepackage{amsmath,amssymb} 
\usepackage{color}

\usepackage[width=122mm,left=12mm,paperwidth=146mm,height=193mm,top=12mm,paperheight=217mm]{geometry}

\usepackage{multirow}
\usepackage{color}
\DeclareMathAlphabet{\pazocal}{OMS}{zplm}{m}{n}

\newcommand\RED[1]{\textcolor{red}{#1}}
\newcommand\BLUE[1]{\textcolor{blue}{#1}}

\definecolor{greyblue}{rgb}{0.1,0.6,0.5}

\definecolor{purple}{rgb}{0.626,0.125,0.941}

\definecolor{pp}{rgb}{0.6,0.0,0.6}

\definecolor{mypink}{rgb}{0.858, 0.188, 0.478}

\definecolor{dodgerblue}{rgb}{0.12, 0.56, 1.0}

\begin{document}
\pagestyle{headings}
\mainmatter
\def\ECCVSubNumber{6775}  

\title{Character Region Attention For Text Spotting} 



\titlerunning{CRAFTS}
%
\author{Youngmin Baek \and Seung Shin \and Jeonghun Baek \and
Sungrae Park \and Junyeop Lee \and Daehyun Nam \and Hwalsuk Lee\thanks{Corresponding author.}}
\authorrunning{Baek et al.}
%
\institute{Clova AI Research, NAVER Corp.\\
\email{\{youngmin.baek,seung.shin,jh.baek,sungrae.park,junyeop.lee,\\daehyun.nam,hwalsuk.lee\}@navercorp.com}}

\maketitle

\begin{abstract}

A scene text spotter is composed of text detection and recognition modules. Many studies have been conducted to unify these modules into an end-to-end trainable model to achieve better performance. A typical architecture places detection and recognition modules into separate branches, and a RoI pooling is commonly used to let the branches share a visual feature. However, there still exists a chance of establishing a more complimentary connection between the modules when adopting recognizer that uses attention-based decoder and detector that represents spatial information of the character regions. This is possible since the two modules share a common sub-task which is to find the location of the character regions. Based on the insight, we construct a tightly coupled single pipeline model. This architecture is formed by utilizing detection outputs in the recognizer and propagating the recognition loss through the detection stage. The use of character score map helps the recognizer attend better to the character center points, and the recognition loss propagation to the detector module enhances the localization of the character regions. Also, a strengthened sharing stage allows feature rectification and boundary localization of arbitrary-shaped text regions. Extensive experiments demonstrate state-of-the-art performance in publicly available straight and curved benchmark dataset.

\end{abstract}

\section{Introduction}

\begin{figure}[t]
  \centering
  \includegraphics*[width=0.9\linewidth, clip=true]{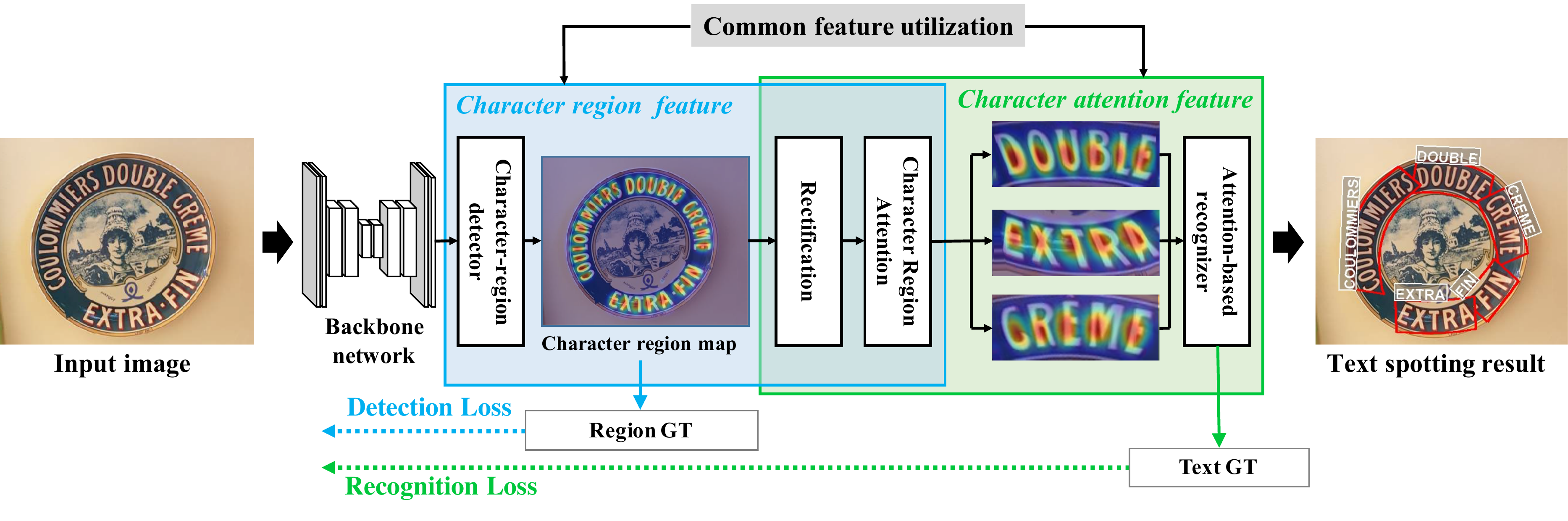}
  \vspace{-5mm}
  \caption{Concept of the proposed method. The character region feature from the detector is used as an input character attention feature to the recognizer. Having a tightly coupled architecture lets the recognition loss flow through the whole network. }
  \label{fig:intro} 
  \vspace{-5mm}
\end{figure}

Scene text spotting, including text detection and recognition, has recently attracted much attention because of its variety of applications in instant translation, image retrieval, and scene parsing. Although existing text detectors and recognizers work well on horizontal texts, it still remains as a challenge when it comes to spotting curved text instances in scene images.

To spot curved texts in an image, a classic method is to cascade existing detection and recognition models to manage text instances on each side. The detectors\cite{lyu2018mask,long2018textsnake,baek2019character} attempt to capture the geometric attributes of curved texts by applying complicated post-processing techniques, and the recognizers apply multi-directional encoding\cite{cheng2018aon} or take rectification modules\cite{shi2018aster,zhan2019esir,gao2018recurrent} to enhance the accuracy of the recognizer on curved texts.

As deep learning advanced, researches have been made to combine detectors and recognizers into a jointly trainable end-to-end network \cite{he2018end,liu2018fots}. Having a unified model not only provides efficiency in the size and speed of the model, but also helps the model learn a shared feature that pulls up the overall performance. To gain benefit from this property, attempts have also been made to handle curved text instances using an end-to-end model\cite{lyu2018mask,qin2019towards,feng2019textdragon,xing2019convolutional}. 
However, most of the existing works only adopt a RoI pooling to share low-level feature layers between detection and recognition branches. In the training phase, instead of training the whole network, only shared feature layers are trained using both detection and recognition losses.

As shown in Fig. \ref{fig:intro}, we propose a novel end-to-end \textit{Character Region Attention For Text Spotting} model, referred to as CRAFTS. Instead of isolating detection and recognition modules in two separate branches, we form a single pipeline by establishing a complimentary connection between the modules. We observe that recognizer \cite{baek2019wrong} using attention-based decoder and detector \cite{baek2019character} encapsulating character spatial information share a common sub-task which is to localize character regions. By tightly integrating the two modules, the outputs from the detection stage helps recognizer attend better to the character center points, and loss propagated from recognizer to detector stage enhances the localization of the character regions. Furthermore, the network is able to maximize the quality of the feature representation used in the common sub-tasks. To best of our knowledge, this is the first end-to-end work that constructs a tightly coupled loss propagation.

The summary of our contribution follows;
(1) We propose an end-to-end network that could detect and recognize arbitrary-shaped texts.
(2) We construct a complementary relationship between the modules by utilizing spatial character information from the detector on the rectification and recognition module.
(3) We establish a single pipeline by propagating the recognition loss throughout all the features in the network.
(4) We achieve the state-of-the-art performances in IC13, IC15, IC19-MLT, and TotalText~\cite{karatzas2013icdar,karatzas2015icdar,nayef2019icdar2019,ch2017total} datasets that contain numerous horizontal, curved and multilingual texts.

\section{Related Work}

{\bf Text detection and recognition methods }
Detection networks use regression based \cite{hu2017wordsup,liao2018textboxes++,liao2018rotation,zhong2019anchor} or segmentation based \cite{deng2018pixellink,long2018textsnake,wang2019shape,xu2019textfield} methods to produce text bounding boxes. Some recent methods like \cite{huang2019mask,liu2019pyramid,zhang2019look} take Mask-RCNN \cite{he2017mask} as the base network and gain advantages from both regression and segmentation methods by employing multi-task learning. In terms of units for text detection, all methods could also be sub-categorized depending on the use of word-level or character-level\cite{hu2017wordsup,baek2019character} predictions. 

Text recognizers typically adopt CNN-based feature extractor and RNN based sequence generator, and are categorized by their sequence generators; connectionist temporal classification (CTC)~\cite{shi2016end} and attention-based sequential decoder~\cite{lee2016recursive,shi2016robust}. Detection model provides information of the text regions, but it is still a challenge for the recognizer to extract useful information in arbitrary-shaped texts. To help recognition networks handle irregular texts, some researches ~\cite{shi2016robust,liu2016star,shi2018aster} utilize spatial transformer network (STN)\cite{jaderberg2015spatial}. Also, the papers ~\cite{gao2018recurrent,zhan2019esir} further extend the use of STN by iterative executing the rectification method. These studies show that running STN recursively helps recognizer extract useful features in extremely curved texts. 
In \cite{liu2018char}, Recurrent RoIWarp Layer was proposed to crop individual characters before recognizing them. The work proves that the task of finding a character region is closely related to the attention mechanism used in the attention-based decoder.

One way to construct a text spotting model is to sequentially place detection and recognition networks. A well known two-staged architecture couples TextBox++\cite{liao2018textboxes++} detector and CRNN\cite{shi2016end} recognizer. With its simplicity, the method achieves favorable results.

\vspace{2mm}
{\bf End-to-end using RNN-based recognizer }
EAA\cite{he2018end} and FOTS\cite{liu2018fots} are end-to-end models based on EAST detector \cite{zhou2017east}. The difference between these two networks lies in the recognizer. The FOTS model uses CTC decoder \cite{shi2016end}, and the EAA model uses attention decoder \cite{shi2016robust}. Both works implement an affine transformation layer to pool the shared feature. The proposed affine transformation works well on horizontal texts, but shows limitations when handling arbitrary-shaped texts. TextNet \cite{sun2018textnet} proposed a spatial-aware text recognizer with perspective-RoI transformation in the feature pooling layer. The network keeps an RNN layer to recognize a sequence of text in the 2D feature map, but due to the lack of expressively of the quadrangles, the network still shows limitations when detecting curved texts.

Qin et al. \cite{qin2019towards} proposed a Mask-RCNN\cite{he2017mask} based end-to-end network. Given the box proposals, features are pooled from the shared layer and the ROI-masking layer is used to filter out the background clutters. The proposed method increases its performance by ensuring attention only in the text region. Busta et al. proposed Deep TextSpotter \cite{busta2017deep} network and extended their work in E2E-MLT \cite{busta2018e2e}. The network is composed of FPN based detector and a CTC-based recognizer. The model predicts multiple languages in an end-to-end manner. 

\vspace{2mm}
{\bf End-to-end using CNN-based recognizer }
Most CNN-based models that recognize texts in character level have advantages when handling arbitrary-shaped texts. MaskTextSpotter \cite{lyu2018mask} is a model that recognizes text using a segmentation approach. Although it has strengths in detecting and recognizing individual characters, it is difficult to train the network since character-level annotations are usually not provided in the public datasets. CharNet \cite{xing2019convolutional} is another segmentation-based method that makes character level predictions. The model is trained in a weakly-supervised manner to overcome the lack of character-level annotations. During training, the method performs iterative character detection to create pseudo-ground-truths.

While segmentation-based recognizers have shown great success, the method suffers when the number of target characters increases. Segmentation based models require more output channels as the number of character sets grow, and this increases memory requirements. The journal version of MaskTextSpotter\cite{liao2019mask} expands the character set to handle multiple languages, but the authors added a RNN-based decoder instead of using their initially proposed CNN-based recognizer. Another limitation of segmentation-based recognizer is the lack of contextual information in the recognition branch. Due to the absence of sequential modeling like RNNs, the accuracy of the model drops under noisy images.
 
TextDragon \cite{feng2019textdragon} is another segmentation-based method that localize and recognize text instances. However, a predicted character segment is not guaranteed to cover a single character region. To solve the issue, the model incorporates CTC to remove overlapping characters. The network shows good detection performance but shows limitations in the recognizer due to the lack of sequential modeling.

\section{Methodology}

\subsection{Overview}

\begin{figure*}[h]
  \begin{center}
    \includegraphics*[width=0.95\linewidth, clip=true]{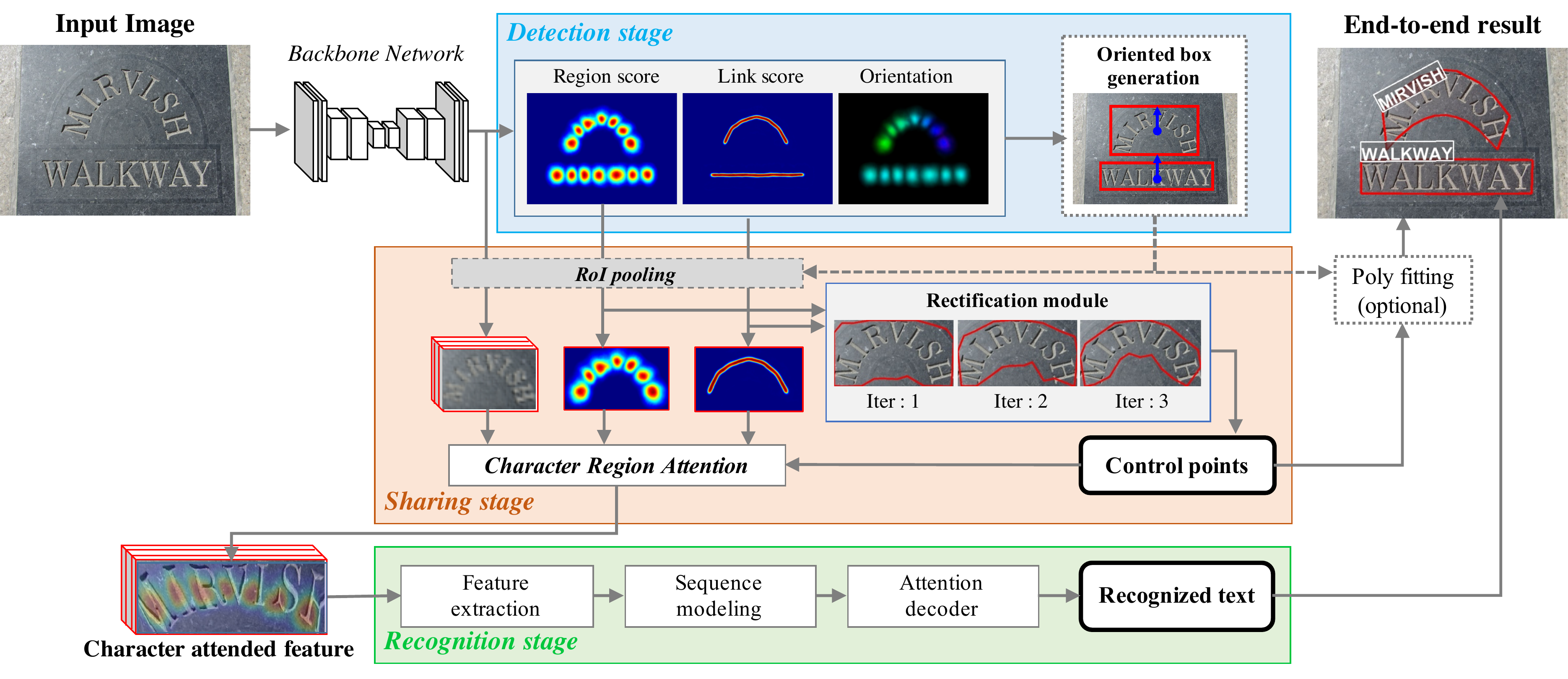}
    \vspace{-1mm}
    \caption{Schematic overview of CRAFTS pipeline.}
\vspace{-10mm}
  \label{fig:architecture} 
  \end{center}
\end{figure*}

Proposed CRAFTS network can be divided into three stages; detection stage, sharing stage, and recognition stage. A detailed pipeline of the network is illustrated in Fig. \ref{fig:architecture}. Detection stage takes an input image and localizes oriented text boxes. Sharing stage then pools backbone high-level features and detector outputs. The pooled features are then rectified using the rectification module, and are concatenated together to form a \textit{character attended feature}. In the recognition stage, attention-based decoder predicts text labels using the \textit{character attended feature}. Finally, a simple post-processing technique is optionally used for better visualization.

\subsection{Detection Stage}
CRAFT detector\cite{baek2019character} is selected as a base network because of its capability of representing semantic information of the character regions. The outputs of the CRAFT network represent center probability of character regions and linkage between them. We contemplate that this character centeredness information can be used to support the attention module in the recognizer since both modules aim to localize the center position of characters. In this work, we make three changes in the original CRAFT model; backbone replacement, link representation, and orientation estimation.

{\bf Backbone replacement } Recent studies show that the use of ResNet50 captures well-defined feature representations of both the detector and the recognizer\cite{long2018scene,baek2019wrong}. We therefore replace the backbone of the network from VGG-16\cite{simonyan2014very} to ResNet50\cite{he2017deep}.  

{\bf Link representation } 
The occurrence of vertical texts is not common in Latin texts, but it is frequently found in East Asian languages like Chinese, Japanese, and Korean.  In this work, a binary center line is used to connect the sequential character regions. This change was made because employing the original affinity maps on vertical texts often produced ill-posed perspective transformation that generated invalid box coordinates.
To generate ground truth linkmap, a line segment with thickness $t$ is drawn between adjacent characters. Here, $t = max((d_1+d_2)/2 * \alpha, 1)$, where $d_1$ and $d_2$ are the diagonal lengths of adjacent character boxes and $\alpha$ is the scaling coefficient. Use of the equation lets the width of the center line proportional to the size of the characters. We set $\alpha$ as 0.1 in our implementation.

{\bf Orientation estimation} 
It is important to obtain the right orientation of text boxes since the recognition stage requires well-defined box coordinates to recognize the text properly. To this end, we add two-channel outputs in the detection stage; channel is used to predict angles of characters along the x-axis, y-axis each. To generate the ground truth of orientation map, the upward angle of the GT character bounding box is represented as $\theta_{box}^{*}$, the channel predicting x-axis has a value of $S_{cos}^{*}(p)=(\cos\theta + 1) \times 0.5$, and the channel predicting y-axis has a value of $S_{sin}^{*}(p)=(\sin\theta + 1)\times 0.5$. The ground truth orientation map is generated by filling the pixels $p$ in the region of the word box with the values of $S_{cos}^*(p)$ and $S_{sin}^*(p)$. The trigonometric function is not directly used to let the channels have the same output range with the region map and the link map; between 0 and 1.

The loss function for orientation map is calculated by Eq. \ref{eq:loss_theta}.
\begin{equation} \label{eq:loss_theta}
L_{\theta} = S_{r}^{*}(p) \cdot (||S_{sin}(p)-S^{*}_{sin}(p)||^{2}_{2} + ||S_{cos}(p)-S^{*}_{cos}(p)||^{2}_{2})
\end{equation}
\noindent where $S_{sin}^{*}(p)$ and $S_{cos}^{*}(p)$ denote the ground truth of text orientation. Here, the character region score $S_{r}(p)$ is used as a weighting factor because it represents the confidence of the character centeredness. By doing this, the orientation loss is calculated only in the positive character regions.

The final objective function in the detection stage $L_{det}$ is defined as,
\begin{equation} \label{eq:loss_d}
L_{det} = L_{r} + L_{l} + \lambda L_{\theta}
\end{equation}
where $L_r$ and $L_l$ denote character region loss and link loss, which are exactly same in \cite{baek2019character}. The $L_{\theta}$ is the orientation loss, and is multiplied with $\lambda$ to control the weight. In our experiment, we set $\lambda$ to 0.1.

\begin{figure}[h]
	\begin{center}
        \includegraphics*[width=0.95\linewidth, clip=true]{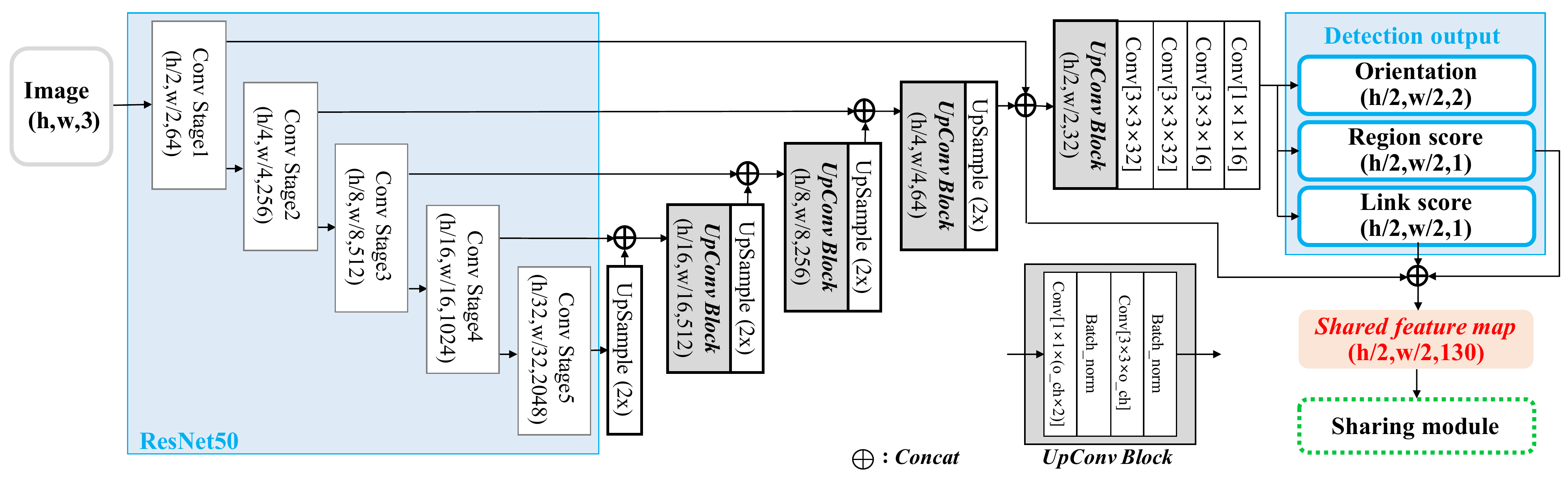}
        \caption{Schematic illustration of the backbone network and the detection head.}
  	    \label{fig:network} 
    \end{center}
    \vspace{-5mm}
\end{figure}

The architecture of the backbone and modified detection head is illustrated in Fig. \ref{fig:network}. The final output of the detector has four channels, each representing \textit{character region map} $S_{r}$, \textit{character link map} $S_{l}$, and two \textit{orientation maps} $S_{sin}, S_{cos}$. 

During inference, we apply the same post-processing as described in \cite{baek2019character} to obtain text bounding boxes. First, by using predefined threshold values, we make binary maps of \textit{character region map} $S_{r}$ and \textit{character link map} $S_{l}$. Then, using the two maps, the text blobs are constructed by using connected components labeling(CCL). The final boxes are obtained by finding a minimum bounding box enclosing each text blob. We additionally determine the orientation of the bounding box by utilizing pixel-wise averaging scheme. As shown in the Eq. \ref{eq:orientation}, the angle of the text box is found by taking the arctangent of accumulated sine and cosine values at the predicted orientation map.
\begin{equation} \label{eq:orientation}
    \theta_{box} = \arctan\left(\frac{\sum(S_{r}(p) \times (S_{sin}(p) - 0.5)}{\sum(S_{r}(p) \times (S_{cos}(p) - 0.5)}\right)
\end{equation}
$\theta_{box}$ denotes orientation of the text box, $S_{cos}$ and $S_{sin}$ are the 2-ch orientation outputs. The same character centerdeness-based weighting scheme that used in the loss calculation is applied to predict the orientation as well.

\subsection{Sharing Stage}
Sharing stage consists of two modules: text rectification module and character region attention(CRA) modules. To rectify arbitrarily-shaped text region, a thin-plate spline (TPS) \cite{shi2018aster} transformation is used. Inspired by the work of \cite{zhan2019esir}, our rectification module incorporates iterative-TPS to acquire a better representation of the text region. By updating the control points attractively, the curved geometry of a text in an image becomes ameliorated. Through empirical studies, we discover that three TPS iterations are sufficient for rectification.

\begin{figure}[h]
  \centering
  \includegraphics*[width=0.9\linewidth, clip=true]{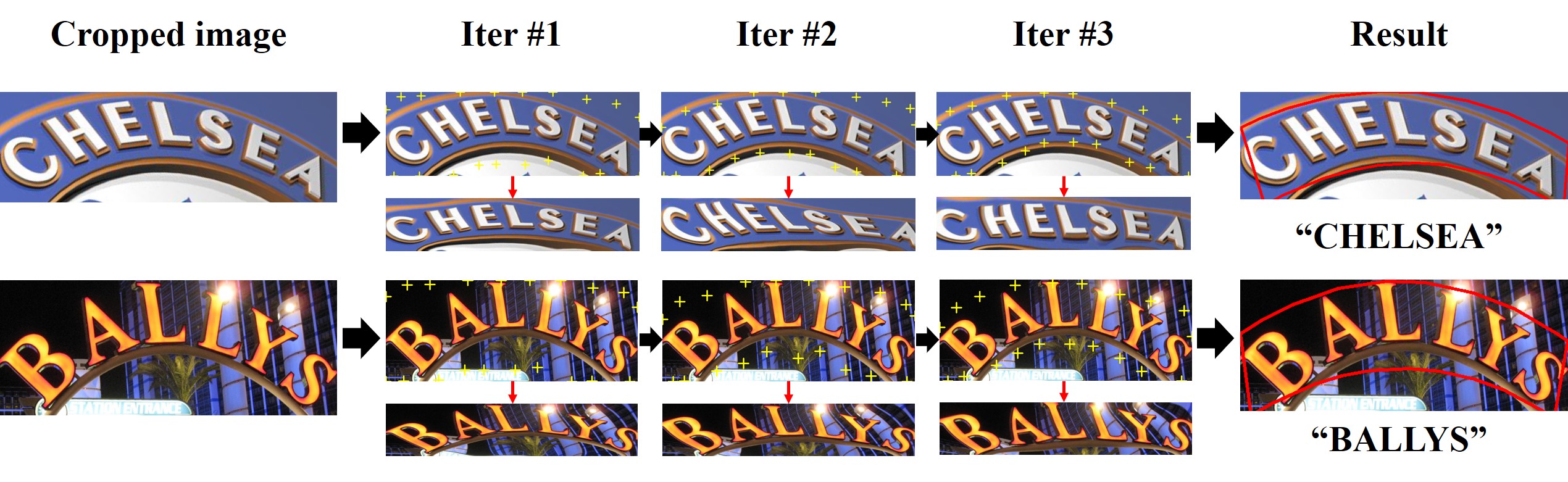}
  \caption{Example of iterative TPS. The middle rows show TPS control points on each iteration, and the bottom row shows rectified images on each stage. The control points are drawn in the image level for better visualization. Actual rectification is done in the feature space. Final result was smoothed using a 2D polynomial.}  
  \label{fig:tps_example}
  \vspace{-5mm}
\end{figure}

Typical TPS module takes an word image as input, but we feed the character region map and link map since they encapsulate geometric information of the text regions. We use twenty control points to tightly cover the curved text region. To use these control points as a detection result, they are transformed to the original input image coordinate. We optionally perform 2D polynomial fitting to smooth the bounding polygon. Examples of iterative-TPS and final smoothed polygon output are shown in Fig. \ref{fig:tps_example}. 

CRA module is the key component that tightly couples detection and recognition modules. By simply concatenating rectified character score map with feature representation, the model establishes following advantages. Creating a link between detector and recognizer allows recognition loss to propagate through detection stage, and this improves the quality of character score map. Also, attaching character region map to the feature helps recognizer attend better to the character regions. Ablation study of using this module will be discussed further in the experiment section.

\subsection{Recognition Stage}
The modules in the recognition stage are formed based on the results reported in \cite{baek2019wrong}. There are three components in the recognition stage: feature extraction, sequence modeling, and prediction. The feature extraction module is made lighter than a solitary recognizer since it takes high-level semantic features as input. 

Detailed architecture of the module is shown in Table \ref{tab:recog_network}. After extracting the features, a bidirectional LSTM is applied for sequence modeling, and attention-based decoder makes a final text prediction.

\begin{table}[h]
\small
\centering
\tabcolsep=0.10cm
\renewcommand*{\arraystretch}{0.7}
\fontsize{7}{7} 
\begin{tabular}[t]{|c|cc|c|}
\hline 
\textbf{Layers} & \multicolumn{2}{c|}{\textbf{Configurations}} & \textbf{Output} \\
\hline 
Input & \multicolumn{2}{c|}{pooled feature} & $64\times 16 \times 130$ \\
\hline 
Block1& \multicolumn{2}{c|}{$\begin{bmatrix}\rm{c:}256,\rm{k:}3\times3\\\rm{c:}256,\rm{k:}3\times3\end{bmatrix}\times 2$} & $64\times16\times256$\\
\hline 
Conv1 & c: $256$ & k: $3\times3$ & $64\times16\times256$\\
\hline
\multirow{2}{*}{MaxPool} & k: $2\times2$ &  & \multirow{2}{*}{$65\times8\times256$}\\
 & s: $1\times2$ & p: $1\times0$ & \\
\hline 
Block2& \multicolumn{2}{c|}{$\begin{bmatrix}\rm{c:}512,\rm{k:}3\times3\\\rm{c:}256,\rm{k:}3\times3\end{bmatrix}\times 5$} & $65\times8\times512$\\
\hline 
Conv2 & c: $512$ & k: $3\times3$ & $65\times8\times512$\\
\hline
Block3& \multicolumn{2}{c|}{$\begin{bmatrix}\rm{c:}512,\rm{k:}3\times3\\\rm{c:}512,\rm{k:}3\times3\end{bmatrix}\times 3$} & $65\times8\times512$\\
\hline 
\multirow{2}{*}{Conv3} & c: $512$ & k: $2\times2$ & \multirow{2}{*}{$65\times4\times512$}\\
 & s: $1\times2$ & p: $1\times0$ & \\
\hline
\multirow{2}{*}{Conv4} & c: $512$ & k: $2\times2$ & \multirow{2}{*}{$65\times3\times512$}\\
 & s: $1\times1$ & p: $0\times0$ & \\
\hline
\multirow{2}{*}{AvgPool} & k: $1\times3$ &  & \multirow{2}{*}{$65\times1\times512$}\\
 & s: $1\times2$ & p: $1\times0$ & \\
\hline 
\end{tabular}
\caption{A simplified ResNet feature extraction module .}
\vspace{-8mm}
\label{tab:recog_network}
\end{table}

\begin{figure}[h]
  \centering
  \includegraphics*[width=0.7\linewidth, clip=true]{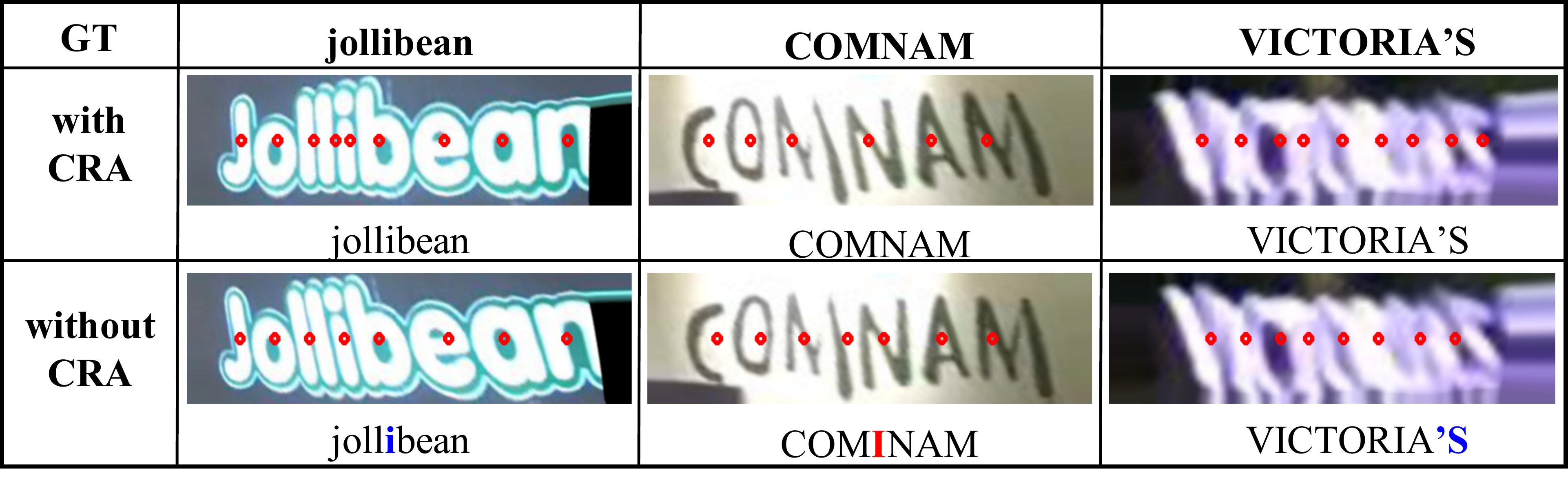}
  \caption{Attention problems with and without Character Region Attention module. The red dots represent attention points of the decoding characters. Missing characters are colored in \BLUE{blue}, and misrecognized characters are colored in \RED{red}. The cropped images are slightly different since generated control points in each rectification modules are inconsistent.}  
  \label{fig:attention_example} 
\end{figure}

At each time step, attention-based recognizer decodes textual information by masking attention outputs to the features. Although attention module works well in most cases, it fails to predict characters when attention points are misaligned or vanished \cite{cheng2017focusing,he2018end}. Fig. \ref{fig:attention_example} shows the effect of using CRA module. Well-placed attention points allow robust text prediction.

The objective function, $L_{reg}$, in the recognition stage is 
\begin{equation} \label{eq:loss_r}
L_{reg}=-\sum_{i}{\log p(Y_{i}|X_{i}})
\end{equation}
 where $p(Y_{i}|X_{i})$ indicates the generation probability of the character sequence, $Y_{i}$, from the cropped feature representation, $X_{i}$ of the $i$-th word box.

\begin{figure}[h]
  \centering
  \includegraphics*[width=0.95\linewidth, clip=true]{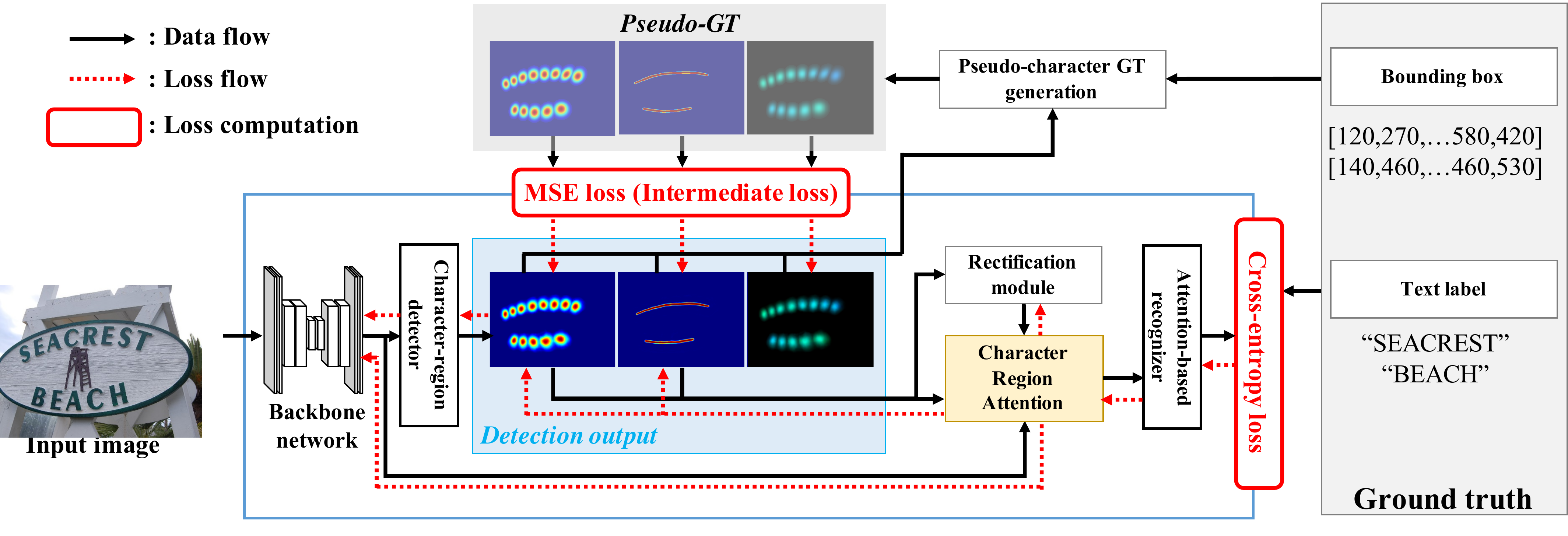}
  \caption{The entire loss flow of CRAFTS model.}
  \vspace{-6mm}
  \label{fig:loss_flow} 
\end{figure}

The final loss, $L$, used for training is composed of detection loss and recognition loss by taking $L=L_{det}+L_{reg}$. The overall flow of the recognition loss is shown in Fig. \ref{fig:loss_flow}. The loss flows through the weights in the recognition stage, and propagates towards detection stage through \textit{Character Region Attention} module. Detection loss on the other hand is used as an intermediate loss, and thus the weights before detection stage are updated using both detection and recognition losses.

\vspace{-6mm}
 \vspace{-6mm}
 
\section{Experiment}

\subsection{Datasets}

\noindent\textbf{English datasets} \textit{IC13}~\cite{karatzas2013icdar} dataset consists of high-resolution images, 229 for training and 233 for testing. A rectangular box is used to annotate word-level text instances. \textit{IC15}~\cite{karatzas2013icdar} consists of 1000 training and 500 testing images. A quadrilateral box is used to annotate word-level text instances. \textit{TotalText}~\cite{ch2017total} has 1255 training and 300 testing images. Unlike IC13 and IC15 datasets, it contains curved text instances and is annotated using polygon points. 

\noindent\textbf{Multi-language dataset} \textit{IC19}~\cite{nayef2019icdar2019} dataset contains 10,000 training and 10,000 testing images. The dataset contains texts in 7 different languages and is annotated using quadrilateral points.

\subsection{Training strategy}
We jointly train both the detector and recognizer in the CRAFTS model. To train the detection stage, we follow the weakly-supervised training method described in \cite{baek2019character}. The recognition loss is calculated by making a batch of randomly sampled cropped word features in each image. Maximum number of words per image is set to 16 to prevent out-of-memory error. Data augmentations in the detector apply techniques like crops, rotations, and color variations. For the recognizer, the corner points of the ground truth boxes are perturbed in a range between 0 to 10\% of the shorter length of the box. 

The model is first trained on the SynthText dataset~\cite{gupta2016synthetic} for 50k iterations, and we further train the network on target datasets. Adam optimizer is used, and \textit{On-line Hard Negative Mining(OHEM)} \cite{shrivastava2016training} is applied to enforce 1:3 ratio of positive and negative pixels in the detection loss. When fine-tuning the model, SynthText dataset is mixed with the ratio of 1:5. We take 94 characters to cover alphabets, numbers, and special characters, and take 4267 characters for the multi-language dataset.

\begin{table*}[t]
  \centering
  \small
  \tabcolsep=0.04cm
  \fontsize{8}{8}\selectfont
  \renewcommand*{\arraystretch}{1.1}
  \begin{tabular}{c||c|c|c|c|c|c||c|c|c|c|c|c||c}
    \hline 
    \rule{0pt}{8.5pt} \multirow{2}{*}{\textbf{Method}} & \multicolumn{3}{c|}{\textbf{IC13}(Det)} & \multicolumn{3}{c||}{\textbf{IC13}(E2E)} & \multicolumn{3}{c|}{\textbf{IC15}(Det)} & \multicolumn{3}{c||}{\textbf{IC15}(E2E)} & \multirow{2}{*}{\textbf{FPS}} \\
    \cline{2-13}
    \rule{0pt}{8.5pt} & \textbf{R} & \textbf{P} & \textbf{H} & \textbf{S} & \textbf{W} & \textbf{G} & \textbf{R} & \textbf{P} & \textbf{H} & \textbf{S} & \textbf{W} & \textbf{G} &\\
    \hline
    \hline

    \rule{0pt}{8.5pt}
    Deep TextSpotter\cite{busta2017deep} & - & - & - & 89 & 86 & 77 & - & - & - & 54 & 51 & 47 & 9\\
    TextBoxes++$^*$\cite{liao2018textboxes++} & 86 & 92 & 89 & 93 & 92 & 85 & 78.5 & 87.8 & 82.9 & 73.3 & 65.8 & 51.9 & -\\
    TextNet$^*$\cite{sun2018textnet} & 89.1 & 93.6 & 91.3 & 89.7 & 88.8 & 82.9 & 80.8 & 85.7 & 83.2 & 78.6 & 74.9 & 60.4 & 2.7\\
    EAA\cite{he2018end} & 89 & 91 & 90 & 91 & 89 & 86 & 86 & 87 & 87 & 82 & 77 & 63 & -\\
    TextDragon\cite{feng2019textdragon} & - & - & - & - & - & - & 83.7 & 92.4 & 87.8 & 82.5 & 78.3 & 65.1 & -\\
    FOTS$^*$\cite{liu2018fots} & - & - & 92.8 & 91.9 & 90.1 & 84.7 & 87.9 & 91.8 & 89.8 & 83.5 & 79.1 & 65.3 & 7.5\\
    Li et al.\cite{li2019towards} & 80.5 & 91.4 & 85.6 & 92.5 & 91.2 & 84.9 & - & - & - & 84.4 & 78.9 & 66.1 & 1.3\\
    Qin et al.\cite{qin2019towards} & - & - & - & - & - & - & 87.9 & 91.6 & 89.7 & \textbf{85.5} & 81.9 & 69.9 & 4.7\\
    CharNet$^*$\cite{xing2019convolutional} & - & - & - & - & - & - & \textbf{90.4} & \textbf{92.6} & \textbf{91.5} & 85.0 & 81.2 & 71.0 & -\\
    MaskTextSpotter$^*$\cite{liao2019mask} & 89.5 & 94.8 & 92.1 & 93.3 & 91.3 & 88.2 & 87.3 & 86.6 & 87.0 & 83.0 & 77.7 & 73.5 & 2.0\\

    \hline
    \hline
    \rule{0pt}{8.5pt} \textbf{CRAFTS(ours)} & \textcolor{black}{\textbf{90.9}} & \textcolor{black}{\textbf{96.1}} & \textcolor{black}{\textbf{93.4}} & \textcolor{black}{\textbf{94.2}} & \textcolor{black}{\textbf{93.8}} & \textcolor{black}{\textbf{92.2}} & 85.3 & 89.0 & 87.1 & 83.1 & \textbf{82.1} & \textbf{74.9} & 5.4\\
    \hline
  \end{tabular}
  \vspace{3mm}
  \caption{Results on horizontal Latin datasets. $^*$ denote the results based on multi-scale tests. R, P, and H refer to recall, precision and H-mean, and S, W, and G indicate strongly-, weakly- and generic-contextualization results, respectively. The best score is highlighted in \textbf{bold}. The evaluation metric of ICDAR 2013 detection task is DetEval, and IoU metric is used for other three cases. FPS is for reference only due to the different experimental environments.}
  \vspace{-6mm}
  \label{tab:result_icdar}
\end{table*}

\vspace{-1mm}

\subsection{Experimental Results}

\noindent\textbf{Horizontal datasets (IC13, IC15)} To target the IC13 benchmark, we take the model trained on the SynthText dataset and perform finetuning on IC13 and IC19 datasets. During inference, we resize the longer side of the input to 1280. The results show significant increase in performance when compared with the previous state-of-the-art works.

The model trained on IC13 dataset is then fine-tuned on the IC15 dataset. During the evaluation process, the input size of the model is set to 2560x1440. Note that we perform generic evaluation without the generic vocabulary set. The quantitative results on IC13 and IC15 datasets are listed in Table. \ref{tab:result_icdar}.

Our method surpasses previous methods in both generic and weakly- contextualization end-to-end tasks, and shows comparable results in other tasks. The generic performance is meaningful because a vocabulary set is not provided in practical scenarios. Note that we get slightly low detection scores on IC15 dataset and also observe low performance in strongly-contextualization results. The relatively low detection performance is obtained mainly due to the granularity difference, and will be discussed further in the later section.

\begin{figure*}[t]
  \centering
  \includegraphics*[width=0.95\linewidth, clip=true]{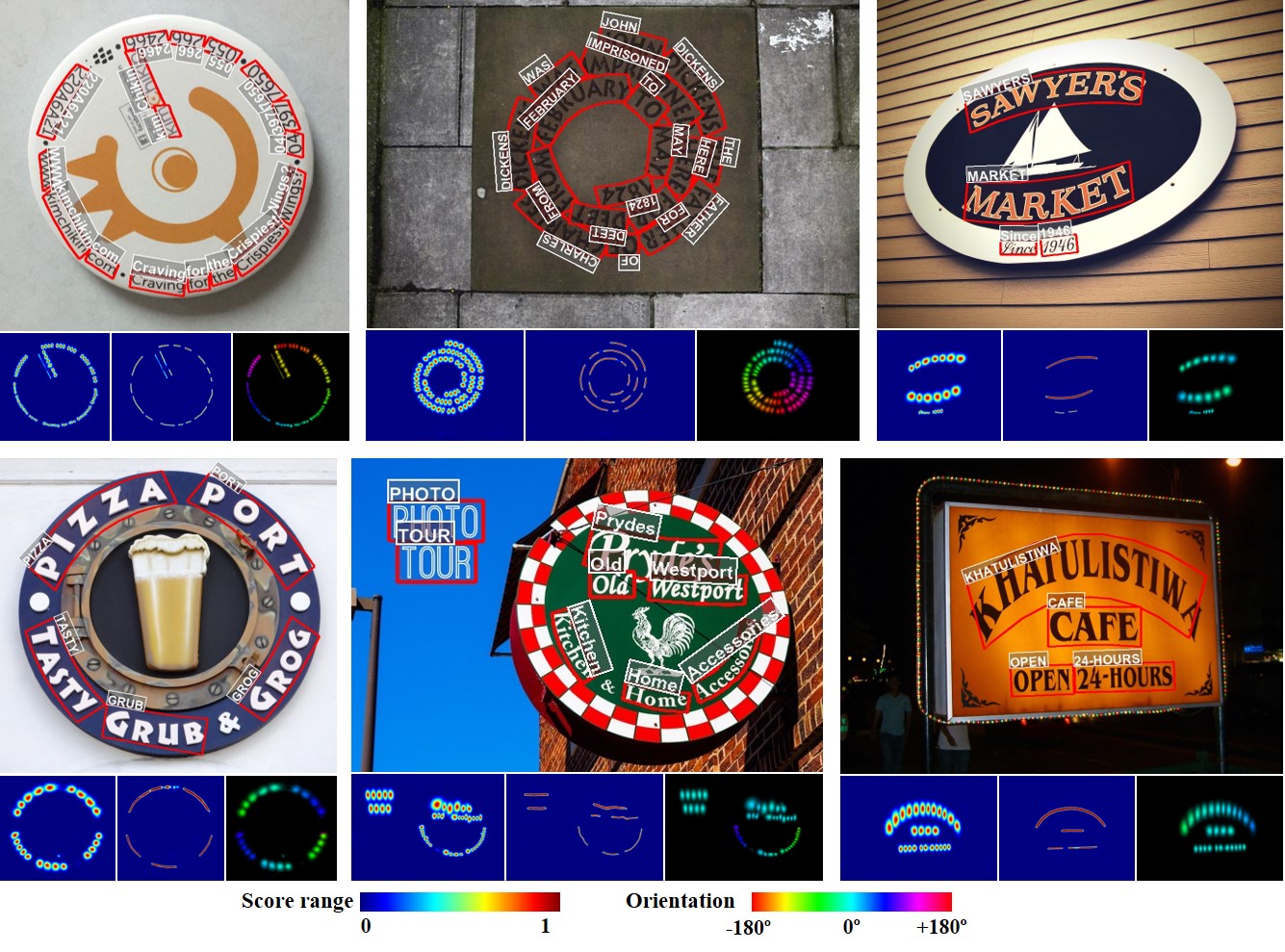}
  \vspace{-3mm}
  \caption{Results on the TotalText dataset. First row: each column shows the input image (top) with its respective region score map, link map, and orientation map.}
  \label{fig:result_curved} 
\end{figure*}

\noindent\textbf{Curved datasets (TotalText)} 
From the model trained on IC13 dataset, we further train the model on TotalText dataset. During inference, we resize the longer side of the input to 1920, and the control points from rectification module are used for detector evaluation. The qualitative results are shown in Fig. \ref{fig:result_curved}. The character region map and the link map are illustrated using a heatmap, and the weighted pixel-wise angle values are visualized in the HSV color space. As it is shown in the figure, the network successfully localizes polygon regions and recognizes characters in the curved text region. Two top-left figures show successful recognition of fully rotated and highly curved text instances.
\vspace{-4mm}
\begin{table}[h]
  \centering
  \tabcolsep=0.1cm
  \fontsize{9.5}{9.5}\selectfont
  \renewcommand*{\arraystretch}{1.0}
  \begin{tabular}{c||c|c|c||c|c|c||c}
    \hline  
    \rule{0pt}{10pt} \multirow{2}{*}{\textbf{Method}} & \multicolumn{3}{c||}{\textbf{Detection}} & \multicolumn{3}{c||}{\textbf{E2E(None)}} & \textbf{E2E(Full)}\\
   \cline{2-8}
    \rule{0pt}{10pt} & \textbf{R} & \textbf{P} & \textbf{H} & \textbf{R} & \textbf{P} & \textbf{H} & \textbf{H}\\
    \hline
    \hline
    TextDragon\cite{feng2019textdragon} & 75.7 & 85.6 & 80.3 & - & - & 48.4 & 74.8\\
    TextNet\cite{sun2018textnet} & 59.4 & 68.2 & 63.5 & 56.4 & 51.9 & 54.0 &  - \\ 
    Li et al.\cite{li2019towards} & 59.8 & 64.8 & 62.2 & - & - & 57.8 & - \\
    MaskTextSpotter\cite{liao2019mask} & 75.4 & 81.8 & 78.5 & - & - & 65.3 & 77.4\\
    CharNet$^*$\cite{xing2019convolutional} & 85.0 & 88.0 & 86.5 & - & - & 69.2 & - \\ 
    Qin et al.$\dagger$\cite{qin2019towards} & 85.0 & 87.8 & 86.4 & - & - & 70.7 & -\\
    \hline
    \rule{0pt}{10pt} \textbf{CRAFTS(ours)} & \textbf{85.4} & \textbf{89.5} & \textbf{87.4} & \textbf{72.2} & \textbf{86.5} & \textbf{78.7} & -\\
    \hline
  \end{tabular}
  \caption{Results on TotalText dataset. None means no lexicon is used for contextualization. The full lexicon contains all words in the test set. $^*$ indicates the multi-scale inference, and $\dagger$ denotes models trained on the private datasets.}
  \label{tab:result_totaltext}
\end{table}
  \vspace{-7mm}

Quantitative results on TotalText dataset are listed in Table. \ref{tab:result_totaltext}. DetEval\cite{ch2017total} evaluates the performance of the detector and modified IC15 evaluation scheme measures the end-to-end performance. Our method outperforms previously reported methods by a large margin. Note that even without the vocabulary set, the end-to-end result significantly exceeds the h-mean score by 8.0\%.

\begin{figure*}[h]
  \centering
  \includegraphics*[width=0.95\linewidth, clip=true]{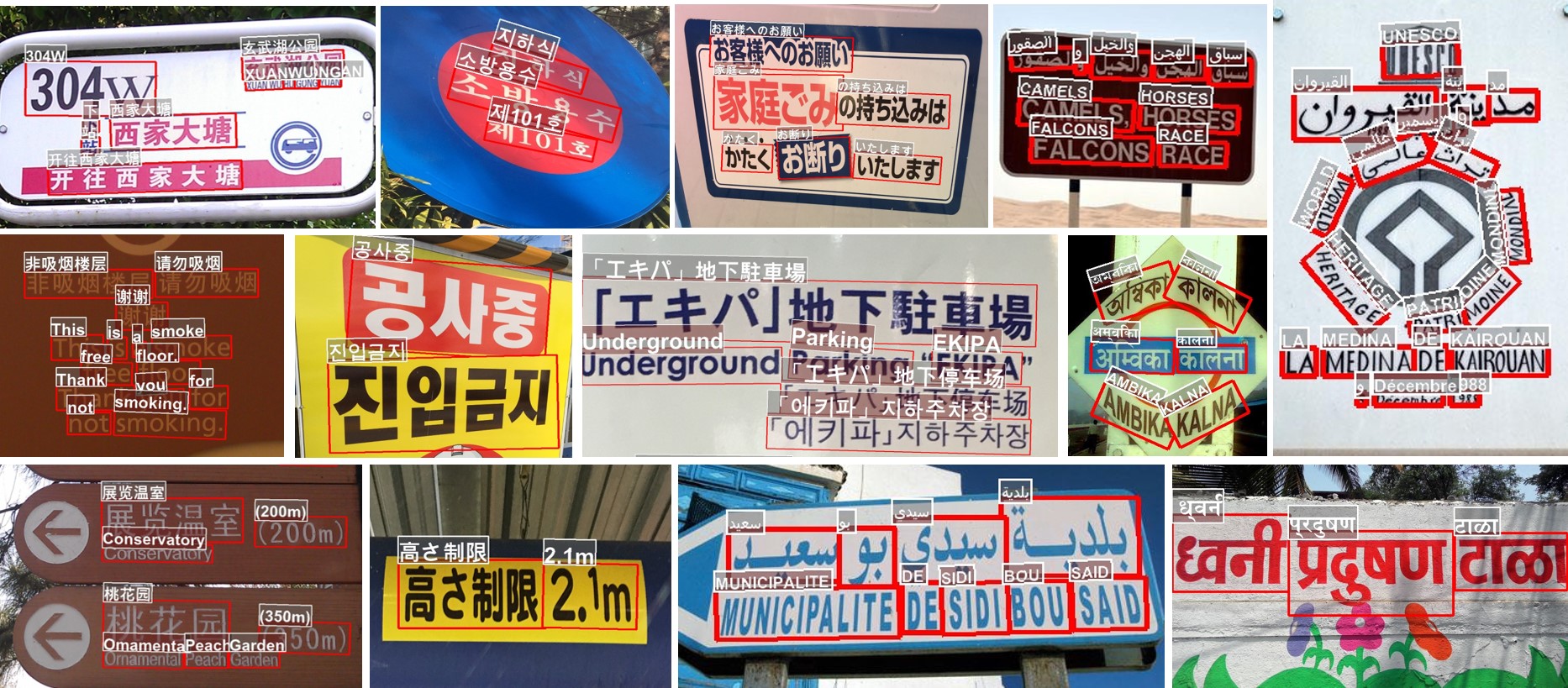}
  \caption{Qualitative results on IC19 MLT dataset.}  
  \label{fig:result_mlt}
  \vspace{-2mm}
\end{figure*}

\noindent\textbf{Multi-language dataset (IC19)} 
Evaluation on multiple languages is performed using IC19-MLT dataset. The output channel in the prediction layer of the recognizer was expanded to 4267 to handle  the characters in Arabic, Latin, Chinese, Japanese, Korean, Bangladesh, and Hindi. However, occurrence of characters in the dataset is not evenly distributed. Among 4267 characters in the training set, 1017 characters occur once in the dataset, and this insufficiency makes it hard for the model to make accurate label predictions. 
To solve class imbalance problem, we first freeze the weights in the detection stage and pretrain the weights in the recognizer with other publicly available multi-language datasets: SynthMLT, ArT, LSVT, ReCTS and RCTW \cite{busta2018e2e,chng2019art,sun2019lsvt,shi2017icdar2017}. We then let the loss flow through the whole network and use IC19 dataset to finetune the model. Since no paper reports performance, we compare our results with E2E-MLT \cite{busta2018e2e,nayef2019icdar2019}.
The samples from the IC19 dataset are shown in Fig. \ref{fig:result_mlt}. We hope our study is set as a baseline for future works on the IC19-MLT benchmark.

\vspace{-4mm}
\begin{table}[h]
  \centering
  \tabcolsep=0.1cm
  \fontsize{9.5}{9.5}\selectfont
  \renewcommand*{\arraystretch}{1.0}
  \begin{tabular}{c||c|c|c||c|c|c}
    \hline  
    \rule{0pt}{10pt} \multirow{2}{*}{\textbf{Method}} & \multicolumn{3}{c||}{\textbf{Detection}} & \multicolumn{3}{c}{\textbf{E2E}}\\
    \cline{2-7}
    \rule{0pt}{10pt} & \textbf{R} & \textbf{P} & \textbf{H} & \textbf{R} & \textbf{P} & \textbf{H}\\
    \hline
    \hline
    E2E-MLT\cite{busta2018e2e} & - & - & - & 20.5 & 37.4 & 26.5\\
    \hline
    \rule{0pt}{10pt} \textbf{CRAFTS(ours)} & \textbf{70.1} & \textbf{81.7} & \textbf{75.5} & \textbf{48.5} & \textbf{72.9} & \textbf{58.2}\\
    \hline
  \end{tabular}
  \caption{Results on IC19-MLT dataset.}
  \vspace{-3mm}
  \label{tab:result_ic19}
\end{table}

\vspace{-5mm}
\subsection{Ablation study}

\noindent\textbf{Attention assisted by \textit{Character Region Attention}} 
In this section, we study how \textit{Character Region Attention(CRA)} affects the performance of the recognizer by training a separate network without CRA.

Table. \ref{tab:ablation_pooling} shows the effect of using CRA on benchmark datasets. 
Without CRA, we observe performance drops on all of the datasets. Especially on the perspective dataset (IC15) and the curved dataset (TotalText), we observe a greater gap when compared with the horizontal dataset (IC13). This implies that feeding character attention information improves the performance of the recognizer when dealing with irregular texts.

\begin{table}[h]
  \small
  \centering
  \tabcolsep=0.1cm
  \fontsize{9}{9}\selectfont
  \renewcommand*{\arraystretch}{1.0}
    \begin{tabular}{c|c|||c|c|c|c||c}
    \hline
    \textbf{Dataset} & \textbf{Type} & \textbf{CRA} & \textbf{R} & \textbf{P} & \textbf{H} & \textbf{Gain} \\ \hline\hline
    \multirow{2}{*}{\textbf{IC13}} & \multirow{2}{*}{\textbf{Horizontal}}
                                        & - & 89.2 & 94.8 & 91.9 & - \\
                                        & & \checkmark & 89.5 & 95.0 & 92.2 & +0.3 \\ \hline
    \multirow{2}{*}{\textbf{IC15}} & \multirow{2}{*}{\textbf{Perspective}}
                                        & - & 64.9 & 84.1 & 73.2 & - \\
                                        & & \checkmark  & 65.9 & 86.7 & 74.9 & +1.7 \\ \hline
    \multirow{2}{*}{\textbf{TotalText}} & \multirow{2}{*}{\textbf{Curved}}
                                        & - & 71.9 & 84.0 & 77.5 & - \\
                                        & & \checkmark & 72.2 & 86.5 & 78.7 & +1.2 \\ \hline
    \end{tabular}
  \caption{The end-to-end performance comparisons of using character attention maps. CRA denotes the use of \textit{Character Region Attention} in recognition stage. R, P, and H refers to Recall, Precision and Hmean values.}
  \vspace{-2mm}
  \label{tab:ablation_pooling}
\end{table}

\noindent\textbf{Recognition loss in the detection stage} The recognition loss flowing through the detection stage affects the quality of the character region map and character link map. It is expected that the recognition loss helps detector localize character regions more explicitly. However, the improvement of character localization is not clearly presented in word-level evaluation. Therefore, in order to show individual character localization ability of the detector, we take advantage of the pseudo-character box generation process in the CRAFT detector. When generating pseudo-ground-truths, supervision network calculates the difference between the number of generated pseudo characters with the number of ground-truth characters in the word transcription. Table. \ref{tab:err_char} shows number of \textit{character error length} on each dataset measured with fully trained networks.

\vspace{-3mm}
\begin{table}[ht]
  \centering
  \tabcolsep=0.1cm
  \fontsize{9}{9}\selectfont
  \renewcommand*{\arraystretch}{1.0}
    \begin{tabular}{c|c||c||c|c||c}
    \hline
    \textbf{Dataset} & \textbf{Total Lengths} & \textbf{SynthText} & \textbf{w.o R-loss} & \textbf{w. R-loss} & \textbf{Diff.} \\ \hline\hline
    ICDAR2015 & 46,107 & 9,324 & 1,251 & 1,147 & -104 (-8.3\%)\\
    TotalText & 53,645 & 16,385 & 5,050 & 4,521 & - 529 (-10.4\%)\\
    \hline
      \end{tabular}
  \caption{Comparison of \textit{character error length} on each dataset with trained networks.}
  \label{tab:err_char}
  \vspace{-7mm}
\end{table}

When training the network on SynthText dataset, \textit{character error length} on each dataset is large. Error decreases further as training is performed on real datasets, but the value drops further by propagating recognition loss to the detection stage. This implies that the use of \textit{Character Region Attention} improves the quality of the localization ability of the detector.

\vspace{2mm}
\noindent\textbf{Importance of orientation estimation} The orientation estimation is important because there are many oriented texts in scene text images. Our pixel-wise averaging scheme is very useful for the recognizer to receive well-defined features. We compare the results of our model when the orientation information is not used. On the IC15 dataset, the performance drops from 74.9\% to 74.1\% (-0.8\%), and on TotalText dataset, the h-mean value drops from 78.7\% to 77.5\% (-1.2\%). The results show that the use of accurate angle information escalates performance on rotated texts.

\subsection{Discussions}
\noindent\textbf{Inference speed} 
Since inference speed varies depending on the input image size, we measure the FPS on different input resolutions, each having a longer side of 960, 1280, 1600, and 2560. The test results give FPS of 9.9, 8.3, 6.8, and 5.4, respectively. For all experiments, we use Nvidia P40 GPU with Intel(R) Xeon(R) CPU. When compared with the 8.6 FPS of the VGG based CRAFT detector\cite{baek2019character}, the ResNet based CRAFTS network achieves higher FPS on the same sized input. Also, directly using the control points from the rectification module alleviates the need of post processing for polygon generation. 

\begin{figure}[h]
  \centering
  \includegraphics*[width=0.85\linewidth, clip=true]{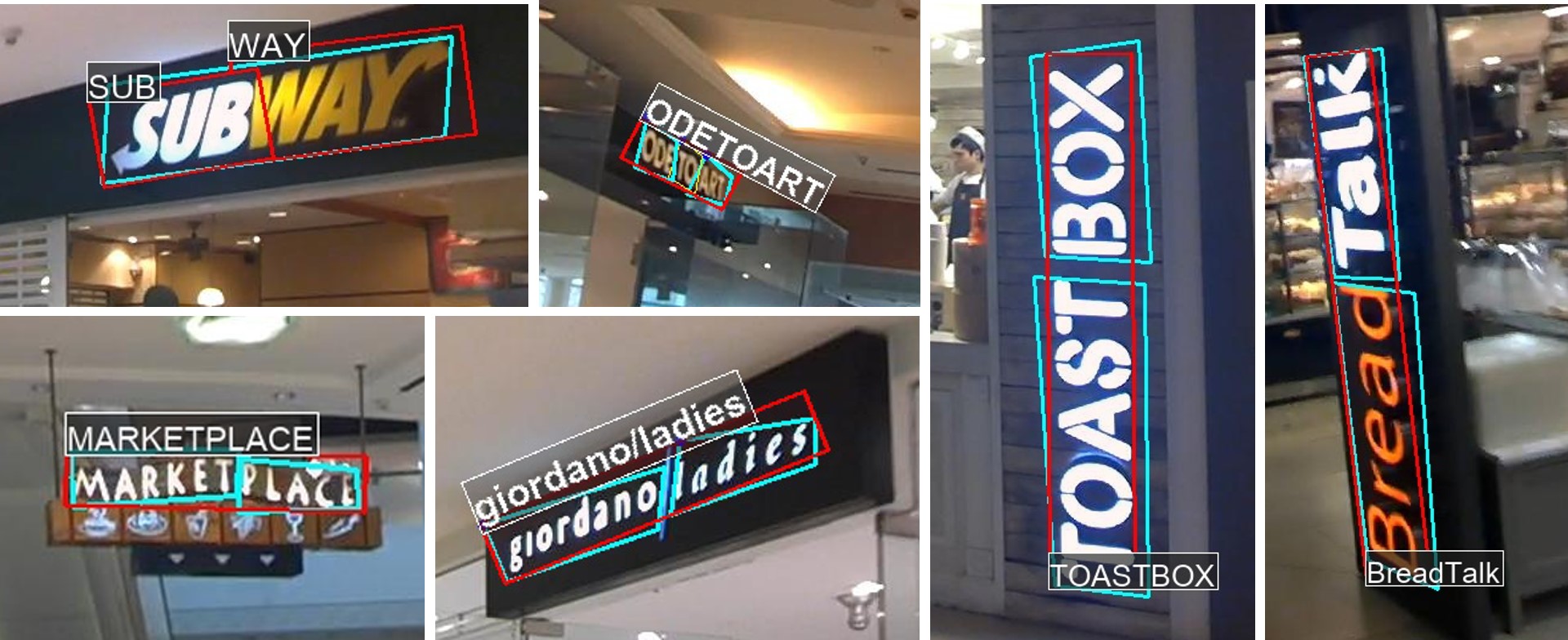}
  \caption{Failure cases in IC15 dataset due to granularity difference. Red boxes denote detection results, cyan boxes denote ground truths.}  
  \label{fig:failurecases} 
  \vspace{-3mm}
 \end{figure}

\noindent\textbf{Granularity difference issue}
We assume that the granularity difference between the ground-truth and prediction box causes relatively low detection performance on the IC15 dataset. 
Character-level segmentation methods tend to generalize character connectivity based on space and color cues, and not capture the whole feature of word instance. For this reason, the outputs do not follow the annotation style of the boxes required by the benchmark. The figure \ref{fig:failurecases} shows the failure cases in the IC15 dataset, which proves that the detection results are marked incorrect while we observe acceptable qualitative results.

\section{Conclusion}
In this paper, we present an end-to-end trainable single pipeline model that tightly couples detection and recognition modules. \textit{Character region attention} in the sharing stage fully exploit \textit{character region map} to help recognizer rectify and attend better to the text regions. Also, we design the recognition loss propagate through detection stage and enhances the character localization ability of the detector. In addition, the rectification module in the sharing stage enables fine localization of curved texts, and obviates the need of developing hand crafted post-processing. The experimental results validate state-of-the-art performance of CRAFTS on various datasets.

\clearpage

\bibliographystyle{splncs04}
\bibliography{mybib}


\end{document}